\pdfoutput=1
\documentclass[11pt]{article}
\usepackage[preprint]{acl}
\usepackage{times}
\usepackage{latexsym}
\usepackage[T1]{fontenc}
\usepackage[utf8]{inputenc}
\usepackage{microtype}
\usepackage{inconsolata}
\usepackage{amsmath,amssymb}
\usepackage{booktabs,tabularx,multirow}
\usepackage{graphicx}
\usepackage{fvextra}
\fvset{fontsize=\footnotesize,breaklines=true,breakanywhere=true}
\usepackage{enumitem}
\usepackage{needspace}
\title{Biomedical Evidence Conflicts: A Dataset and Benchmark for Multi-Source Question Answering}
\author{
  \textbf{Jia Zhang}$^{1}$,
  \textbf{Tengfei Ma}$^{1*}$,
  \textbf{Chengrui Xiang}$^{1}$,
  \textbf{Xiaofen Ouyang}$^{2}$, \\
  \textbf{Jiarui Lu}$^{1}$,
  \textbf{Daojian Zeng}$^{2*}$,
  \textbf{Xiangxiang Zeng}$^{1*}$ \\
  $^{1}$College of Computer Science and Electronic Engineering, Hunan University \\
  $^{2}$College of Information Science and Engineering, Hunan Normal University
}
\begin{document}
\maketitle
\begingroup
\renewcommand\thefootnote{}\footnote{$^{*}$Correspondence to: Daojian Zeng \texttt{<zengdj916@163.com>}, Tengfei Ma \texttt{<tfma@hnu.edu.cn>}, Xiangxiang Zeng \texttt{<xzeng@hnu.edu.cn>}.}
\addtocounter{footnote}{-1}
\endgroup
\begin{abstract}
Navigating biomedical research requires synthesizing primary studies that frequently yield divergent conclusions, yet existing multi-source benchmarks largely assume consistent evidence and overlook reasoning reliability under conflict. To systematically evaluate how reliably large language models (LLMs) reason over conflicting scientific evidence, we construct a large-scale structured dataset of 88,617 question--evidence instances across 296,348 PubMed records, annotated with four fine-grained evidence states, candidate answers, and granular support and attack links. Through expert review and LLM-based quality filtering, we curate a benchmark of 3,602 questions with verified labels and canonical references. Evaluating eight LLMs across evidence recognition, answering behavior, and reference agreement, we find that models systematically underestimate conflict severity and general reasoning outweighs biomedical specialization. While explicit policy guidance improves surface-level behavior and abstention, substantive conflicts remain a severe bottleneck, with reference agreement reaching only 4.10\% to 33.85\%. An automated diagnostic analysis reveals that 69.16\% of failures stem from a structural refusal to commit to a principal conclusion, exposing an entrenched model bias toward hedging when faced with genuine scientific disagreement.
\end{abstract}
\section{Introduction}
Large language models (LLMs) are increasingly deployed as scientific research assistants and clinical synthesis tools, where reliability depends on reconciling primary studies that frequently yield divergent conclusions. Yet existing multi-source question answering (QA) benchmarks largely assume consistent or complementary evidence, evaluating \emph{evidence aggregation}---a regime where frontier LLMs already approach saturation (85\%--95\% pass rates in our evaluation). What remains critically under-tested is \textbf{epistemic reasoning reliability under genuine scientific conflict}: determining whether competing studies reflect compatible findings, reconcilable study conditions, missing evidentiary links, or unreconciled disagreement that requires weighing methodological strength rather than retreating into passive hedging~\citep{han-etal-2026-evidence,cattan2025dragged,yuan-etal-2026-benchmarking}.

Peer-reviewed biomedical literature offers an ideal, high-stakes testbed for this challenge. Unlike open-domain web retrieval benchmarks---where contradictions frequently arise from misinformation, outdated pages, or noisy provenance~\citep{cattan2025dragged,yuan-etal-2026-benchmarking}---divergent findings across PubMed studies reflect structured biological and methodological variation (e.g., cell models, patient populations, or assay endpoints) alongside explicit hierarchies of evidentiary weight (e.g., randomized clinical trials versus preliminary proxy assays). Prior biomedical benchmarks, however, focus on single-document classification (PubMedQA; \citealp{jin-etal-2019-pubmedqa}) or small-scale pairwise contradiction detection (BioConflict; \citealp{kirubakaran-gagnier-2026-bioconflict}), leaving question-centered multi-source conflict reasoning unmeasured at scale.

\begin{table*}[t]
\centering\fontsize{8}{9.5}\selectfont\setlength{\tabcolsep}{4pt}
\begin{tabularx}{\textwidth}{p{1.55cm}p{3.0cm}XX}
\toprule
Evidence state & Question & Findings in the supplied abstracts & Expected answer pattern \\
\midrule
Compatible & How is interferon associated with leukopenia? & Interferon-based treatment is associated with leukopenia; adding matrine reduces this adverse-event risk relative to interferon alone. & Integrate the adverse-event association and the compatible treatment comparison. \\
\addlinespace
Apparent conflict & In human aortic vascular smooth muscle cells, how does oxLDL affect cytochrome c release? & A vascular smooth-muscle study reports increased release; a monocytic-cell study reports attenuated translocation. & Give the increased-release conclusion for the specified cell model and cite the applicable source. \\
\addlinespace
Substantive conflict & What is the relationship between ANP and renin? & Rat kidney and cortical-slice studies report stimulation, while other rat studies report inhibition. The reviewed reference selects inhibition. & State one evidence-based principal conclusion, with material opposing findings retained as limitations. \\
\addlinespace
Insufficient evidence & How does Na$^+$ affect prolactin? & The abstracts study prolactin's effects on sodium levels or transport, rather than sodium's effects on prolactin. & State that the asked relationship is not established and identify the directional evidence gap. \\
\bottomrule
\end{tabularx}
\caption{Four evidence states, illustrated with reviewed items. Findings are paraphrased from the fixed evidence packs. The apparent-conflict question is conditioned for QA; its category is defined on the original broad question. Full item identifiers and source links appear in Appendix~\ref{app:examples}.}
\label{tab:examples}
\end{table*}

\begin{figure*}[t]
\centering
\includegraphics[width=\textwidth]{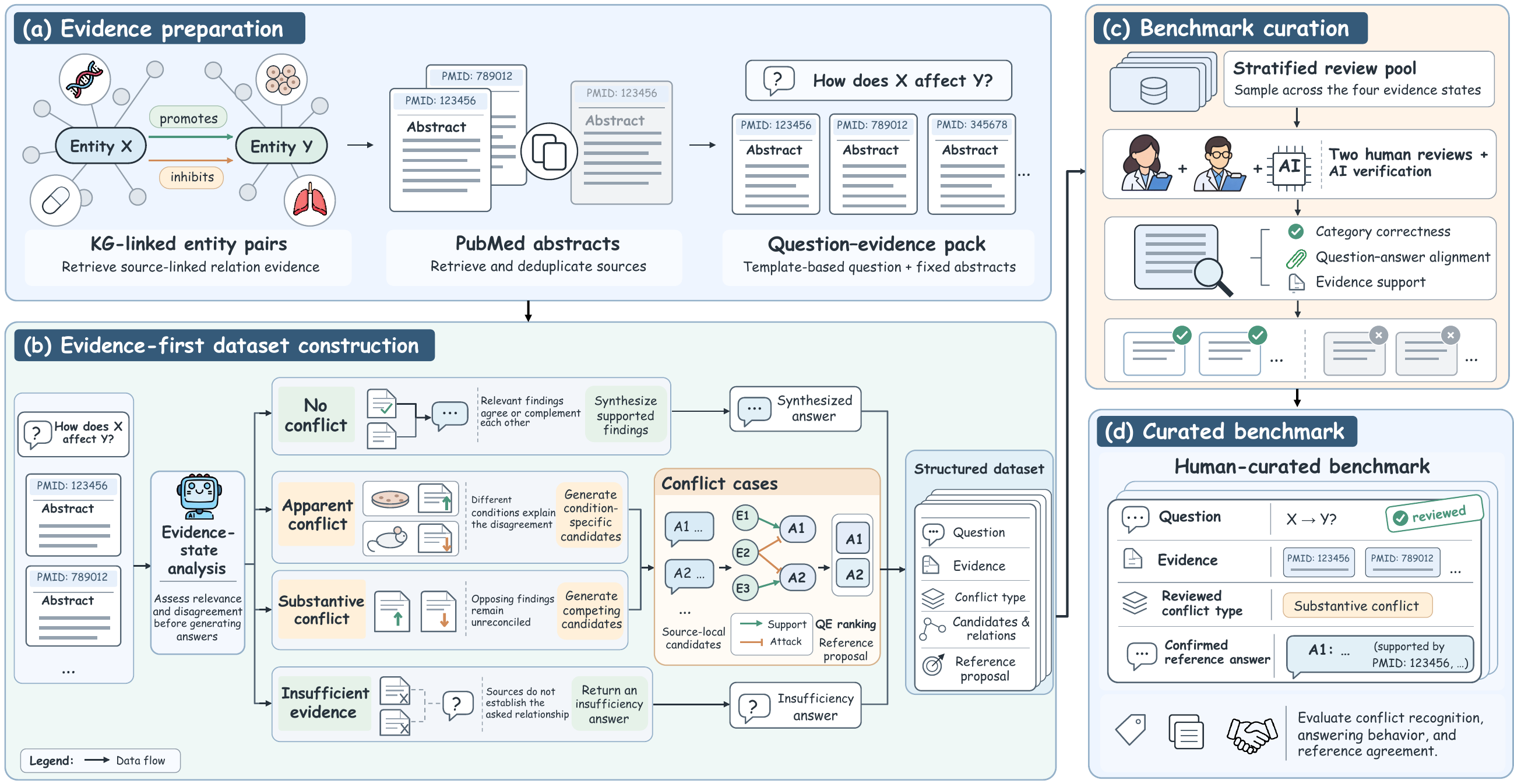}
\caption{Dataset construction and benchmark curation. The large dataset provides structured multi-source evidence annotations; the human-curated benchmark supplies the evaluation set used in all reported model experiments.}
\label{fig:construction}
\end{figure*}

To fill this gap, we introduce a large-scale structured dataset of 88,617 question--evidence instances (across 296,348 PubMed records) and a rigorously verified benchmark of 3,602 curated questions grounded in fixed PubMed abstract packs (Figure~\ref{fig:construction}). Rather than treating conflict as binary, we organize multi-source evidence into four operational states (Table~\ref{tab:examples}): \textbf{compatible} findings calling for integration; \textbf{apparent conflicts} explainable by study conditions; \textbf{substantive conflicts} where competing conclusions remain unreconciled and demand evidence-based adjudication; and \textbf{insufficient evidence} where the abstracts fail to establish the asked relationship. By holding evidence packs fixed, we isolate reasoning reliability from retrieval noise and decouple evaluation along three orthogonal axes: \textbf{(i) conflict-type recognition}, \textbf{(ii) evidence-grounded answering behavior}, and \textbf{(iii) principal-conclusion agreement} with expert-reviewed references. This structure probes three progressive capabilities: distinguishing true conflict from conditional divergence or missing evidence, adhering to scoped grounding and abstention when warranted, and adjudicating unreconciled conflicts through evidentiary weighing.

Evaluating eight open-source and proprietary LLMs exposes a pronounced \emph{illusion of competence} hidden by conventional aggregation benchmarks. Across tasks, models systematically underestimate conflict severity, and general-purpose reasoning and abstention outweigh biomedical specialization. While frontier API models excel on compatible, scope-conditioned apparent, and insufficient items---and accurately comprehend entities and endpoints---all models suffer a severe \textbf{substantive-conflict cliff}, with principal-conclusion agreement plunging to \textbf{4.10\%--33.85\%} (Table~\ref{tab:main}). Notably, even under an oracle \emph{category-assisted} setting that supplies both the ground-truth evidence state and explicit handling rules, gains in surface behavioral compliance fail to translate into reliable scientific judgment and instead degrade grounding and internal consistency. Automated failure diagnosis further reveals that \textbf{69.16\% of failures stem from a structural refusal to commit to a principal conclusion} rather than reading miscomprehension, while stronger frontier models shift toward evidence-supported conclusions that diverge from expert prioritization (Figure~\ref{fig:substantive_diagnosis}).

Our contributions are threefold:
\begin{enumerate}[leftmargin=*,itemsep=3pt]
\item \textbf{Large-Scale Structured Conflict Dataset:} We construct 88,617 question--evidence instances across 296,348 distinct PubMed records and 14 entity-type pair categories, annotated with four evidence states, candidate answers, and granular support/attack links.
\item \textbf{Human-Curated Evaluation Benchmark:} Through dual-reviewer verification and LLM-based quality filtering, we curate 3,602 benchmark questions with verified states, scope-conditioned apparent-conflict queries, and evidence-grounded canonical references.
\item \textbf{Systematic Evaluation and Failure Diagnosis:} We evaluate eight diverse LLMs across recognition, grounded answering, and reference agreement, providing fine-grained criterion and error analyses that pinpoint why current models fail to reliably adjudicate genuine scientific conflicts.
\end{enumerate}

\section{Related Work}
\label{sec:related}
\paragraph{Conflicting evidence in QA.}
Recent multi-source QA benchmarks examine conflict recognition and category-appropriate generation over web documents: DRAGged into Conflicts evaluates pipeline and category-assisted settings~\citep{cattan2025dragged}, while ConfRAG separates answer clustering, answer coverage, and reason coverage~\citep{yuan-etal-2026-benchmarking}. Our decoupled protocol shares the motivation that recognizing disagreement, grounding responses, and matching a principal conclusion are distinct capabilities. However, whereas open-domain web conflicts often reflect noisy provenance and reward balanced presentation, our substantive-conflict policy stress-tests evidence-based adjudication over peer-reviewed scientific studies.

\paragraph{Biomedical conflict and context.}
BioConflict assesses contradiction detection, contextual-variable extraction, and consensus synthesis across 250 paper pairs~\citep{kirubakaran-gagnier-2026-bioconflict}, and \citet{han-etal-2026-evidence} analyze conflicting evidence and document-order effects in biomedical QA. Our benchmark complements these studies with a large-scale, question-centered multi-abstract resource that explicitly models evidence sufficiency alongside broad and scope-conditioned queries (Table~\ref{tab:benchmark-comparison}).

\paragraph{Grounding and evidence aggregation.}
FaithEval probes unanswerable, inconsistent, and counterfactual contexts~\citep{ming2025faitheval}; ALCE separates answer correctness from citation quality~\citep{gao-etal-2023-enabling}; and RAGTruth annotates unsupported claims~\citep{niu-etal-2024-ragtruth}. These works motivate decoupling behavioral faithfulness from principal-conclusion agreement. Quantitative bipolar argumentation provides a structured aggregation formalism~\citep{zhu2025argrag}, which we adapt to nominate reference candidates prior to human review.

\begin{table*}[t]
\centering
\scriptsize
\setlength{\tabcolsep}{2pt}
\renewcommand{\arraystretch}{1.05}
\newcommand{\featureyes}{\(\checkmark\)}
\newcommand{\featureno}{\(\times\)}
\begin{tabular*}{\textwidth}{@{\extracolsep{\fill}}llccccl@{}}
\toprule
\textbf{Dataset} & \textbf{Size}
& \shortstack{\textbf{Multiple}\\\textbf{sources}}
& \shortstack{\textbf{Conflict}\\\textbf{types}}
& \shortstack{\textbf{Evidence}\\\textbf{stance}}
& \shortstack{\textbf{Free-form}\\\textbf{task}}
& \shortstack[l]{\textbf{Question}\\\textbf{origin}} \\
\midrule
PubMedQA {\citeyearpar{jin-etal-2019-pubmedqa}}
& 273,518 instances
& \featureno & \featureno & \featureno & \featureno
& Article titles \\
BioASQ 13b {\citeyearpar{nentidis-etal-2025-bioasq13b}}
& 5,729 questions
& \featureyes & \featureno & \featureno & \featureyes
& Expert-written \\
\midrule
ManConCorpus {\citeyearpar{alamri-stevenson-2016-manconcorpus}}
& 259 abstracts
& \featureyes & \featureno & \featureyes & \featureno
& Systematic reviews \\
HealthContradict {\citeyearpar{zhang-etal-2026-healthcontradict}}
& 920 instances
& \featureyes & \featureno & \featureyes & \featureno
& TREC queries \\
BioConflict {\citeyearpar{kirubakaran-gagnier-2026-bioconflict}}
& 250 abstract pairs
& \featureyes & \featureyes & \featureno & \featureyes
& \textemdash \\
\midrule
\textbf{Ours}
& \textbf{88,617 instances}
& \featureyes & \featureyes & \featureyes & \featureyes
& KG-based templates \\
\bottomrule
\end{tabular*}
\caption{Comparison of biomedical QA and evidence-conflict datasets. Evidence stance marks support or opposition at the document or claim level. A free-form task evaluates a generated answer, summary, or synthesis. PubMedQA includes 1,000 expert-labeled, 61,249 unlabeled, and 211,269 automatically generated instances. Our dataset pairs each question with 2--4 PubMed abstracts and links sources to candidate answers. The evaluation benchmark comprises 3,602 reviewed questions.}
\label{tab:benchmark-comparison}
\end{table*}

\section{Benchmark Construction and Evaluation}
\label{sec:dataset}
Figure~\ref{fig:construction} summarizes our end-to-end pipeline: we first construct a large-scale structured dataset of graph-linked PubMed evidence packs (\S\ref{sec:structured-dataset}), then curate a human-verified evaluation benchmark (\S\ref{sec:benchmark}), and finally evaluate models under a decoupled three-task protocol (\S\ref{sec:protocol}).

\subsection{Large-Scale Structured Dataset}
\label{sec:structured-dataset}
\paragraph{Question--evidence packs.}
We mine the iKraph full graph~\citep{zhang2025ikraph} for entity pairs supported by both positive and negative relations, requiring at least two distinct PubMed IDs per polarity with relation extraction scores of at least 0.7. Selecting two IDs per polarity, retrieving their abstracts, and deduplicating yields fixed source packs. Instantiating entity-type questions (e.g., how a compound affects a gene or how a gene is associated with a disease) produces 107,854 eligible question--evidence packs. We use the graph to identify candidate entity pairs and source PMIDs, and determine the evidence state directly from each question and its retrieved abstracts.

\paragraph{Evidence-state annotation.}
We adapt the Biomni A1 agent~\citep{huang2025biomni} with its biomedical system prompt and DeepSeek-V4.1-Flash as the backbone. As shown in Figure~\ref{fig:construction}(b), the agent first identifies relevant abstracts and assigns one of four evidence states to the original question before any answers are generated (Table~\ref{tab:examples}):
\begin{description}[leftmargin=0pt,itemsep=2pt]
\item[Compatible (no conflict).] Relevant findings agree or complement one another without a supported opposing claim.
\item[Apparent conflict.] Findings appear opposed, but reported study conditions explain why they apply in different settings; methodological differences alone do not explain the disagreement.
\item[Substantive conflict.] Relevant findings oppose one another, and the supplied abstracts do not reconcile the disagreement.
\item[Insufficient evidence.] The abstracts do not establish the asked relationship (e.g., addressing a neighboring endpoint or the reverse direction).
\end{description}
\begin{table}[!t]
\centering\fontsize{8}{9.5}\selectfont
\begin{tabular}{lrr}
\toprule Evidence state & Dataset & Benchmark \\
\midrule
Compatible & 32,453 & 1,440 \\
Apparent & 44,769 & 901 \\
Substantive & 8,890 & 780 \\
Insufficient & 2,505 & 481 \\
\midrule
Total instances & 88,617 & 3,602 \\
\midrule
Distinct PubMed records & 296,348 & 13,744 \\
Entity-type pair categories & 14 & 13 \\
Mean abstracts per instance & 3.82 & 3.85 \\
\bottomrule
\end{tabular}
\caption{The structured dataset and its human-curated benchmark. Dataset categories are automatically annotated; benchmark categories are confirmed during curation. ``Compatible'' denotes the no-conflict class.}
\label{tab:dataset}
\end{table}

To ensure biomedical precision during state assignment, reported null findings count as evidence about the relationship, and the annotator distinguishes association from causation as well as distinct outcomes such as gene expression versus gene activity.

\paragraph{Candidate generation and reference proposal.}
The assigned evidence state routes answer generation: compatible items synthesize supported findings, insufficient items return a fixed abstention response, and both conflict states generate one candidate per relevant abstract with supporting quotations and reported study conditions. For conflict items, the agent annotates each abstract as supporting, challenging, or neutral toward each candidate within its claim scope, alongside inter-abstract relations (using reasoning-enabled calls for state/relation labeling and non-reasoning calls for candidate generation). To nominate a pre-review reference candidate, we rank candidates using quadratic energy (QE) semantics in a quantitative bipolar argumentation graph adapted from \citet{zhu2025argrag} (Appendix~\ref{app:qe}). We use QE to nominate initial reference candidates for review, while final benchmark inclusion and canonical references are confirmed directly from the abstract text.

\begin{figure}[!t]
\centering
\captionsetup{skip=3pt}
\includegraphics[width=0.84\linewidth]{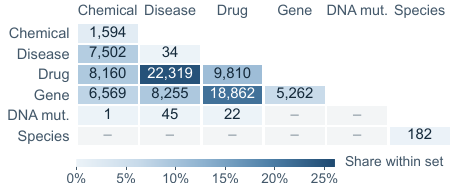}\par
\caption{Entity-type pairs in the full dataset.}
\label{fig:entity-pairs-dataset}
\vspace{3pt}
\includegraphics[width=0.84\linewidth]{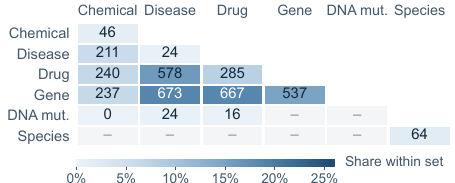}\par
\caption{Entity-type pairs in the benchmark.}
\label{fig:entity-pairs-benchmark}
\end{figure}

\paragraph{Dataset statistics.}
The pipeline yields 88,617 structured instances across 296,348 distinct PubMed records and 230,778 answer candidates (mean 3.82 abstracts per instance). Each record stores the question, graph provenance, abstracts, evidence state, candidates with support/attack links, and proposed reference (Table~\ref{tab:dataset}; Figure~\ref{fig:entity-pairs-dataset} shows the 14 entity-type pairs, led by drug--disease and drug--gene).

\subsection{Human-Curated Benchmark}
\label{sec:benchmark}
We derive our evaluation benchmark by verifying labels and references directly against the supplied abstracts (Figure~\ref{fig:construction}c--d).

\paragraph{Stratified review and verification.}
We sample 5,000 instances stratified by proposed state: 1,500 each for compatible, apparent conflict, and substantive conflict, and 500 for insufficient evidence. Two independent domain reviewers evaluate three binary criteria: category correctness, question--answer alignment, and complete evidence support. Inter-reviewer agreement is high across all axes---88.36\%, 88.18\%, and 88.14\%, respectively (Appendix~\ref{app:dataset-stats}). Retaining only items confirmed by both reviewers across all three criteria yields 4,165 instances, which then pass an LLM-based quality-filtering gate under the same criteria to remove residual boundary ambiguities, resulting in 3,602 curated questions.

\paragraph{Evaluation questions and adjudication policy.}
Each curated item pairs the original broad question $q$ and fixed abstracts $E$ with a verified state $c$ and canonical reference $r$. For apparent conflict, we construct a scope-conditioned QA question $q'$ incorporating the study condition from the confirmed reference without revealing the outcome (e.g., specifying the cell model in Table~\ref{tab:examples}), while conflict-type recognition always uses the unconditioned $q$.

For substantive-conflict items, reviewers select a canonical reference based on evidentiary weight reported within the abstracts---such as prioritizing randomized clinical trials or direct endpoints over preliminary or proxy findings (e.g., \texttt{KGPUB\_062102}; Appendix~\ref{app:wording})---retaining items where one conclusion is clearly better supported. Enforcing this explicit adjudication policy directly tests whether LLMs can weigh competing methodological evidence rather than defaulting to passive equivocation.

\paragraph{Benchmark statistics.}
The 3,602 curated questions span 13 entity-type pairs and 13,744 distinct PubMed records, averaging 3.85 abstracts and 929 words per pack (Table~\ref{tab:dataset}; Figure~\ref{fig:entity-pairs-benchmark}). Grouping questions by shared PubMed records forms 3,487 connected clusters used for bootstrap uncertainty estimation. All model evaluations below use this curated benchmark.
\subsection{Decoupled Evaluation Protocol}
\label{sec:protocol}
Each benchmark item is a tuple $(q, E, c, q', r)$, where models observe only source titles, abstracts, and local identifiers; canonical references and curation metadata are withheld. Rather than collapsing performance into a single opaque QA score, we decouple evaluation into three orthogonal tasks covering conflict recognition, evidence-grounded answering behavior, and principal-conclusion agreement.

\paragraph{Task~1: Conflict-type recognition.}
Given the unconditioned pair $(q, E)$, the model predicts one of the four evidence states in an independent call separate from QA, preventing scope hints in $q'$ from leaking into recognition. We report overall accuracy, four-class Macro-F1, and per-class recall.

\paragraph{Task~2: Evidence-grounded answering behavior.}
Given $(q', E)$, the model generates four structured fields: a main conclusion, supporting source IDs, a justification, and limitations (expected patterns in Table~\ref{tab:examples}). A response passes only if it cites valid source IDs and satisfies five checks: \emph{required action} (executing the state-appropriate behavior, e.g., an evidence-justified choice under substantive conflict), \emph{conclusion form} (committing to one stance rather than unranked hedging), \emph{grounding} (claims and preference rationales supported by cited abstracts), \emph{coverage} (addressing key relevant evidence and counterevidence), and \emph{internal consistency} (stance alignment across conclusion, justification, and limitations). The judge sees the question, abstracts, target state, policy, and output, while remaining blind to $r$.

\paragraph{Task~3: Principal-conclusion reference agreement.}
To separate behavioral formatting from scientific judgment, a judge evaluates semantic agreement with $r$ across four checks on the main conclusion and limitations: \emph{entity and endpoint} (matching core entities and biological outcome), \emph{principal relation} (matching direction, polarity, and causal scope), \emph{resolved conclusion} (committing to one explicit position), and \emph{no internal reversal} (limitations not negating the main conclusion). Because Task~2 independently audits coverage and grounding, a response can match $r$ in Task~3 yet fail Task~2 by omitting counterevidence. Full rubrics appear in Appendix~\ref{app:rubrics}.
\section{Experiments and Diagnostic Analysis}
\label{sec:experiments}
We structure our empirical investigation from broad cross-task patterns to deep bottleneck diagnosis: we detail the evaluation setup (\S\ref{sec:setup}), analyze performance across tasks, evidence states, model families, and policy guidance (\S\ref{sec:main-results}), dissect criterion-level performance on substantive conflicts (\S\ref{sec:fine-grained-analysis}), and attribute root causes across 5,376 failing responses (\S\ref{sec:failure-diagnosis}).

\begin{table*}[t]
\centering\fontsize{8}{9.5}\selectfont
\setlength{\tabcolsep}{2.7pt}
\renewcommand{\arraystretch}{1.06}
\begin{tabular}{@{}l@{\hspace{9pt}}rr@{\hspace{9pt}}rrrr@{\hspace{9pt}}rrrr@{}}
\toprule
& \multicolumn{2}{c}{\textbf{Task~1}} & \multicolumn{4}{c}{\textbf{Task~2: Behavior}} & \multicolumn{4}{c}{\textbf{Task~3: Agreement}} \\
\cmidrule(lr){2-3}\cmidrule(lr){4-7}\cmidrule(lr){8-11}
& \multicolumn{2}{c}{Recognition} & \multicolumn{2}{c}{Direct QA} & \multicolumn{2}{c}{Category-assisted} & \multicolumn{2}{c}{Direct QA} & \multicolumn{2}{c}{Category-assisted} \\
\cmidrule(lr){2-3}\cmidrule(lr){4-5}\cmidrule(lr){6-7}\cmidrule(lr){8-9}\cmidrule(lr){10-11}
Model & Acc. & Macro-F1 & Acc. & Macro & Acc. & Macro & Acc. & Macro & Acc. & Macro \\
\midrule
\multicolumn{11}{@{}l}{\textit{Local general-purpose models}} \\
\midrule[0.35pt]
Qwen3.5-9B & \makebox[28pt][r]{58.86} & \makebox[28pt][r]{53.95} & \makebox[28pt][r]{64.46} & \makebox[28pt][r]{61.56} & \makebox[28pt][r]{67.57} & \makebox[28pt][r]{66.92} & \makebox[28pt][r]{69.02} & \makebox[28pt][r]{65.46} & \makebox[28pt][r]{69.91} & \makebox[28pt][r]{68.80} \\
Qwen3.6-35B-A3B & \makebox[28pt][r]{62.08} & \makebox[28pt][r]{52.65} & \makebox[28pt][r]{68.02} & \makebox[28pt][r]{65.06} & \makebox[28pt][r]{70.02} & \makebox[28pt][r]{69.09} & \makebox[28pt][r]{70.38} & \makebox[28pt][r]{66.94} & \makebox[28pt][r]{69.13} & \makebox[28pt][r]{67.40} \\
Gemma-3-27B & \makebox[28pt][r]{50.06} & \makebox[28pt][r]{49.16} & \makebox[28pt][r]{51.25} & \makebox[28pt][r]{44.69} & \makebox[28pt][r]{52.11} & \makebox[28pt][r]{48.61} & \makebox[28pt][r]{62.60} & \makebox[28pt][r]{54.36} & \makebox[28pt][r]{60.88} & \makebox[28pt][r]{55.29} \\
\midrule
\multicolumn{11}{@{}l}{\textit{Biomedical specialist}} \\
\midrule[0.35pt]
MedGemma-27B & \makebox[28pt][r]{57.44} & \makebox[28pt][r]{53.66} & \makebox[28pt][r]{54.86} & \makebox[28pt][r]{49.15} & \makebox[28pt][r]{54.16} & \makebox[28pt][r]{49.87} & \makebox[28pt][r]{65.07} & \makebox[28pt][r]{57.77} & \makebox[28pt][r]{63.83} & \makebox[28pt][r]{57.52} \\
\midrule
\multicolumn{11}{@{}l}{\textit{API-served general-purpose models}} \\
\midrule[0.35pt]
DeepSeek-V4.1-Flash & \makebox[28pt][r]{61.74} & \makebox[28pt][r]{58.14} & \makebox[28pt][r]{72.71} & \makebox[28pt][r]{71.37} & \makebox[28pt][r]{76.18} & \makebox[28pt][r]{74.81} & \makebox[28pt][r]{74.71} & \makebox[28pt][r]{72.60} & \makebox[28pt][r]{75.04} & \makebox[28pt][r]{73.21} \\
MiniMax-M3.1-Flash & \makebox[28pt][r]{75.12} & \makebox[28pt][r]{72.13} & \makebox[28pt][r]{\textbf{74.43}} & \makebox[28pt][r]{\textbf{72.07}} & \makebox[28pt][r]{79.15} & \makebox[28pt][r]{77.88} & \makebox[28pt][r]{76.07} & \makebox[28pt][r]{73.38} & \makebox[28pt][r]{77.57} & \makebox[28pt][r]{75.40} \\
GPT-6.1 Terra & \makebox[28pt][r]{\textbf{78.98}} & \makebox[28pt][r]{\textbf{77.63}} & \makebox[28pt][r]{70.93} & \makebox[28pt][r]{69.46} & \makebox[28pt][r]{77.73} & \makebox[28pt][r]{76.77} & \makebox[28pt][r]{76.29} & \makebox[28pt][r]{73.97} & \makebox[28pt][r]{79.15} & \makebox[28pt][r]{76.96} \\
Gemini 3.6 Flash & \makebox[28pt][r]{78.43} & \makebox[28pt][r]{77.35} & \makebox[28pt][r]{72.27} & \makebox[28pt][r]{70.57} & \makebox[28pt][r]{\textbf{81.59}} & \makebox[28pt][r]{\textbf{81.59}} & \makebox[28pt][r]{\textbf{76.51}} & \makebox[28pt][r]{\textbf{74.54}} & \makebox[28pt][r]{\textbf{80.82}} & \makebox[28pt][r]{\textbf{79.73}} \\
\bottomrule
\end{tabular}
\caption{Overall performance on 3,602 benchmark questions (percent). Task~1 predicts the evidence category from the broad question and source pack. For Task~2 and Task~3, direct QA uses the question and evidence; category-assisted QA also supplies the reviewed category and handling policy. Acc. is accuracy, and Macro is the unweighted mean of the four class pass rates. Bold marks the best score in each column.}
\label{tab:main}
\end{table*}

\subsection{Experimental Setup}
\label{sec:setup}
\paragraph{Models and inference.}
We evaluate eight LLMs: four local models---Qwen3.5-9B, Qwen3.6-35B-A3B, Gemma-3-27B, and the biomedical specialist MedGemma-27B---and four API models---DeepSeek-V4.1-Flash, MiniMax-M3.1-Flash, GPT-6.1 Terra, and Gemini 3.6 Flash (abbreviated as DeepSeek/MiniMax Flash in figures). All models share identical fixed source packs, retaining the first valid answer per item (exact API identifiers and inference details in Appendix~\ref{app:models}).

\begin{figure*}[t]
\centering
\includegraphics[width=0.85\textwidth]{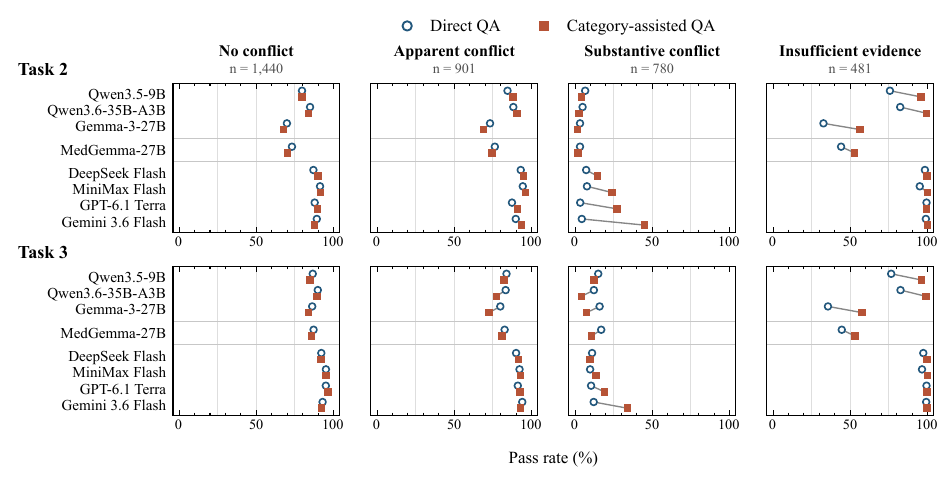}
\caption{Pass rates by evidence state for Task~2 (evidence-grounded behavior) and Task~3 (reference agreement). Connected markers compare direct and category-assisted QA for the same model. Each panel uses the same percentage scale. Counts denote questions per category.}
\label{fig:category_pass_rates}
\end{figure*}

\paragraph{Prompting, judging, and controls.}
We compare \emph{Direct QA} (answering from abstracts without Task~1 predictions) against \emph{Category-assisted QA} (adding the reviewed evidence state for $q$ and handling rule while withholding references and rankings; Task~1 uses the direct setting). Tasks~2 and~3 use DeepSeek-V4.1-Flash as the primary judge; to guard against judge or self-preference bias, we conduct a blind cross-family audit with Gemini 3.6 Flash (Appendix~\ref{app:cross-judge}), alongside prompt-scaffolding (\emph{Policy-only}, \emph{Extract-then-answer}; Appendix~\ref{app:prompt-strategies}) and wording controls (Appendix~\ref{app:wording}) on Qwen3.6-35B-A3B and 3,487-cluster bootstrap intervals (Appendix~\ref{app:classresults}).
\raggedbottom
\begin{figure*}[!t]
\centering
\includegraphics[width=.80\textwidth]{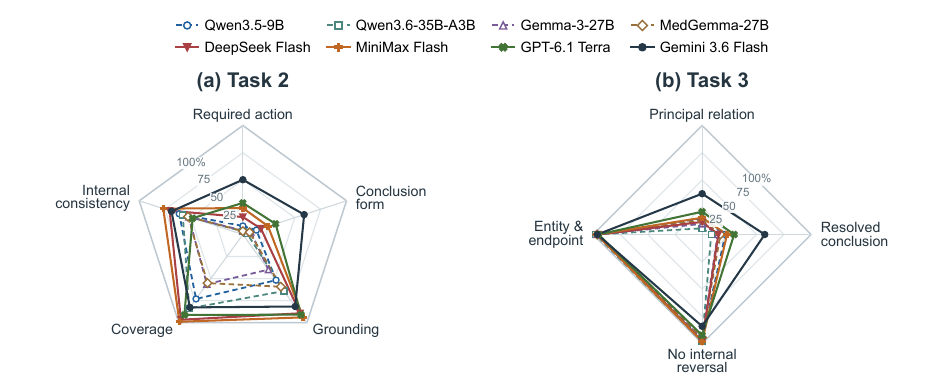}
\caption{Criterion pass rates for category-assisted QA on substantive conflicts (780 questions per model). Panels show the five Task~2 checks and four Task~3 checks (criteria defined in \S\ref{sec:protocol}) on the same 0--100\% scale.}
\label{fig:qa-criterion-radars}
\end{figure*}

\subsection{Main Results Across Tasks, States, and Models}
\label{sec:main-results}
Table~\ref{tab:main} reports overall scores across all 3,602 questions, and Figure~\ref{fig:category_pass_rates} (full table in Appendix~\ref{app:classresults}) breaks down Task~2 and Task~3 across the four evidence states.

\paragraph{Models systematically underestimate conflict severity.}
In Task~1 (Table~\ref{tab:main}; per-class summary in Table~\ref{tab:task1-class}, Appendix~\ref{app:classresults}), models reliably identify compatible evidence (80.5\% recall) and missing links (76.7\%), but frequently downgrade conflicts---misclassifying 29.7\% of apparent conflicts as compatible and 34.9\% of substantive conflicts as apparent.

\paragraph{Substantive conflict creates a sharp performance cliff.}
Across evidence states (Figure~\ref{fig:category_pass_rates}), frontier API models routinely exceed 85\%--95\% on compatible, scope-conditioned apparent, and insufficient items. By contrast, every model collapses on \emph{substantive conflicts}, falling below 50\% on Task~2 and to 4.10\%--33.85\% on Category-assisted Task~3.

\begin{figure*}[t]\centering
\includegraphics[width=.80\textwidth]{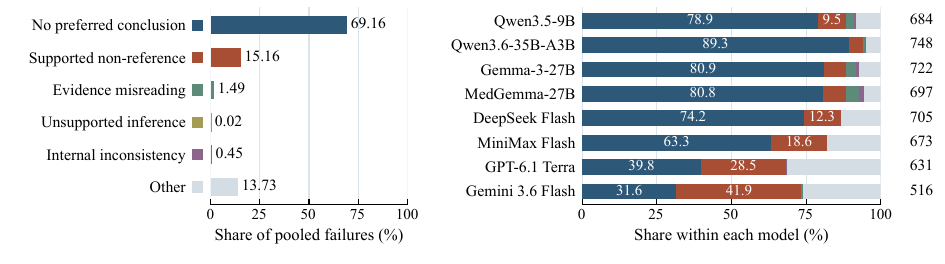}
\caption{Primary diagnostic labels for the 5,376 valid answers that failed Task~3 on substantive conflicts under category-assisted QA. The left panel pools failures across models. The right panel shows within-model proportions and failure counts. Supported non-reference choices are conclusions supported by a supplied source that differ from the reviewed reference.}
\label{fig:substantive_diagnosis}
\end{figure*}

\paragraph{General reasoning and alignment outweigh biomedical specialization.}
Domain tuning lifts MedGemma-27B over its base model Gemma-3-27B on conflict recognition (+7.4 percentage points, pp) and apparent-conflict agreement (+8.3 pp), yet it still trails the 3$\times$-smaller general-purpose Qwen3.5-9B overall (Table~\ref{tab:main}) and shares Gemma's weak abstention on \emph{insufficient evidence} (53\%--56\% vs.\ over 95\% for Qwen and API models; Table~\ref{tab:classfull}).

\paragraph{Explicit policy guidance improves compliance, not scientific judgment.}
Supplying the ground-truth state and answering rule improves Task~2 across all models (Table~\ref{tab:main}), notably lifting average \emph{insufficient-evidence} pass rates from 78.3\% to 87.9\%. On \emph{substantive conflicts}, however, Task~2 rises by +10.0 pp while Task~3 reference agreement barely shifts (+0.9 pp), and supplementary extraction and wording controls fail to close this gap (Appendices~\ref{app:prompt-strategies} and~\ref{app:wording}).

\subsection{Criterion-Level Analysis on Substantive Conflicts}
\label{sec:fine-grained-analysis}
To pinpoint why substantive conflicts remain unsolved under explicit policy guidance, we examine criterion-level pass rates across Task~2 and Task~3 (Figure~\ref{fig:qa-criterion-radars}; full breakdown in Appendix~\ref{app:task2-breakdown}).

\paragraph{Models comprehend the abstracts accurately, but fail to adjudicate.}
All eight models surpass 95\% on entity/endpoint matching and maintain strong grounding and coverage, confirming accurate reading of individual studies. Failure concentrates strictly on decisive resolution---averaging 17.1\% on \emph{required action} and 20.4\% on \emph{conclusion form}, with even Gemini 3.6 Flash matching the reviewed principal relation on only 37.4\% of items.

\paragraph{Pressuring commitment induces a compliance--faithfulness trade-off.}
Moving from Direct to Category-assisted QA more than doubles \emph{required action} (6.3\% $\rightarrow$ 17.1\%) and \emph{conclusion form} (10.0\% $\rightarrow$ 20.4\%), yet reduces \emph{grounding} ($-3.8$ pp) and \emph{internal consistency} ($-6.0$ pp; Appendix~\ref{app:task2-breakdown}), as models pressured to take a stand invoke unsupported rationales or reverse their main conclusion under limitations.
\subsection{Failure Diagnosis and Case Studies}
\label{sec:failure-diagnosis}
To examine how models fail when adjudication breaks down, we apply automated diagnostic attribution to all 5,376 valid Category-assisted substantive-conflict responses failing Task~3 (Figure~\ref{fig:substantive_diagnosis}; per-model breakdown in Appendix~\ref{app:diagnosis}).

\paragraph{Entrenched hedging dominates, shifting to divergent weighing in frontier models.}
Reading errors are negligible (\emph{evidence misunderstanding} accounts for 1.49\%). Instead, \emph{no preferred principal conclusion} dominates at 69.16\% of failures (ranking first in seven of eight models), exposing an entrenched bias toward hedging. As capability scales, failure modes shift rather than disappear: \emph{evidence-supported non-reference choices} rise sharply to 28.53\% in GPT-6.1 Terra and 41.86\% in Gemini 3.6 Flash, showing that stronger models commit more often yet weigh competing evidence differently from reviewers.

\paragraph{Case studies.}
Concrete examples illustrate this contrast (Appendices~\ref{app:diagnosis}--\ref{app:wording}). On \texttt{KGPUB\_073908} (\emph{MTHFR} 677TT and colorectal cancer risk), Qwen3.6-35B-A3B merely lists opposing studies without resolving, whereas Gemini 3.6 Flash commits to reduced risk grounded in a supplied meta-analysis. When preliminary evidence conflicts with a randomized trial (\texttt{KGPUB\_062102}), models often relegate the trial comparison to limitations rather than adjudicating.

\begin{table}[t]
\centering\footnotesize
\setlength{\tabcolsep}{3pt}
\begin{tabular}{@{}lrrrrr@{}}
\toprule
 & & \multicolumn{2}{c}{Task~2 (\%)} & \multicolumn{2}{c}{Task~3 (\%)} \\
\cmidrule(lr){3-4}\cmidrule(l){5-6}
Evidence state & $n$ & Orig. & Gemini & Orig. & Gemini \\
\midrule
Compatible & 96 & 87.50 & 83.33 & 92.71 & 93.75 \\
Apparent conflict & 96 & 91.67 & 89.58 & 88.54 & 94.79 \\
Substantive conflict & 144 & 10.42 & 19.44 & 14.58 & 23.61 \\
Insufficient evidence & 64 & 85.94 & 85.94 & 85.94 & 85.94 \\
\bottomrule
\end{tabular}
\caption{Cross-judge audit pass rates on the 400-item stratified sample (pooling all eight models and both settings). Orig.\ denotes the primary DeepSeek-V4.1-Flash judge; Gemini denotes independent blind grading by Gemini 3.6 Flash.}
\label{tab:cross-judge-states}
\end{table}

\paragraph{Cross-judge robustness.}
A blind audit of 400 stratified responses using Gemini 3.6 Flash achieves 88.25\% agreement ($\kappa=0.752$) on Task~2 and 90.50\% ($\kappa=0.792$) on Task~3 with the primary DeepSeek-V4.1-Flash judge, independently reproducing the substantive-conflict collapse (Table~\ref{tab:cross-judge-states}; full audit in Appendix~\ref{app:cross-judge}).
\section{Conclusion}
We present a large-scale structured dataset of 88,617 question--evidence instances and a human-curated benchmark of 3,602 questions to evaluate LLM reasoning reliability under multi-source biomedical evidence conflict. Decoupling recognition, grounded answering behavior, and principal-conclusion agreement reveals systematic conflict underestimation, stronger gains from general reasoning than domain specialization, and a severe substantive-conflict cliff (4.10\%--33.85\% reference agreement). Explicit category-assisted guidance improves surface compliance but barely shifts scientific judgment and triggers a compliance--faithfulness trade-off, while 69.16\% of failures stem from entrenched hedging and stronger models shift toward uncalibrated evidentiary weighing. Beyond biomedical QA, our findings demonstrate that scaling and prompting alone cannot resolve genuine scientific disagreement, underscoring the need for argumentation-aware alignment and principled evidence adjudication in trustworthy AI research assistants.
\flushbottom
\section*{Limitations}
While our benchmark provides a rigorous testbed for multi-source biomedical knowledge conflicts, several scope boundaries remain and point to concrete future directions.
First, our resource is constructed from fixed packs of graph-linked PubMed abstracts rather than unrestricted web retrieval. While this design isolates multi-source reasoning from retrieval noise, it limits evaluation to concise abstracts rather than full-length articles containing finer experimental protocols and statistical details.
Second, our canonical reference answers represent expert-confirmed targets under a reproducible consensus policy; in highly contentious substantive conflicts, alternative interpretations or supplementary conclusions can also carry partial evidentiary support.
Finally, our study focuses on systematically benchmarking and diagnosing conflict-reasoning failures in existing LLMs rather than developing new training or inference-time mitigation methods. Extending this framework to full-text literature and designing evidence-weighing architectures, argumentation-aware alignment, and multi-turn interactive adjudication to actively improve models' ability to weigh competing scientific evidence remain important directions for future work.

\bibliography{bibliography}
\appendix
\section{Additional Dataset Statistics}\label{app:dataset-stats}
The complete structured dataset contains 88,617 instances, 85,667 distinct normalized question strings, 13,633 distinct typed entities, and 230,778 answer candidates. Instances are identified by entity pair and evidence pack, so different instances may share a question string. Its evidence consists of 338,633 source occurrences and 296,348 distinct PubMed records. Four abstracts are available for 75,712 instances, three for 9,975, and two for 2,930. Source packs average 913 words. The dataset covers 14 entity-type pair categories; the curated benchmark covers 13. All review-pool and benchmark identifiers are verified against the complete dataset.

The review pool contains 5,000 items. Inter-reviewer agreement on the three binary confirmation decisions is 4,418/5,000 (88.36\%) for category correctness, 4,409/5,000 (88.18\%) for question--answer alignment, and 4,407/5,000 (88.14\%) for answer support. Selection requires positive confirmation on all axes, followed by the LLM-based quality-filtering gate.

The retained graph entity-type pair counts are: disease--gene (673), drug--gene (667), disease--drug (578), gene--gene (537), drug--drug (285), chemical--drug (240), chemical--gene (237), chemical--disease (211), species--species (64), chemical--chemical (46), disease--disease (24), DNA mutation--disease (24), and DNA mutation--drug (16). Evidence packs have four sources for 3,173 questions, three for 321, and two for 108. Their median length is 944 words.

\section{Illustrative Items and Source Traceability}
\label{app:examples}
Table~\ref{tab:examples} uses four retained items:
\begin{itemize}[leftmargin=*,itemsep=3pt]
\item \textbf{Compatible:} \texttt{KGPUB\_015431}. Interferon--leukopenia; illustrative sources include PMIDs \href{https://pubmed.ncbi.nlm.nih.gov/22104564/}{22104564}, \href{https://pubmed.ncbi.nlm.nih.gov/25824075/}{25824075}, and \href{https://pubmed.ncbi.nlm.nih.gov/28382770/}{28382770}.
\item \textbf{Apparent conflict:} \texttt{KGPUB\_000120}. The broad question asks how oxLDL affects cytochrome c. The QA question specifies human aortic vascular smooth muscle cells and cytochrome c release. PMID \href{https://pubmed.ncbi.nlm.nih.gov/14551153/}{14551153} supplies the increased-release finding; PMID \href{https://pubmed.ncbi.nlm.nih.gov/17890680/}{17890680} describes attenuated translocation in monocytic cells.
\item \textbf{Substantive conflict:} \texttt{KGPUB\_067067}. ANP--renin; PMIDs \href{https://pubmed.ncbi.nlm.nih.gov/2856728/}{2856728} and \href{https://pubmed.ncbi.nlm.nih.gov/3015641/}{3015641} report stimulation, while \href{https://pubmed.ncbi.nlm.nih.gov/2990872/}{2990872} supplies the selected inhibition reference. The model receives all four abstracts, including PMID \href{https://pubmed.ncbi.nlm.nih.gov/2953525/}{2953525}.
\item \textbf{Insufficient evidence:} \texttt{KGPUB\_007734}. Na$^+$--prolactin; PMIDs \href{https://pubmed.ncbi.nlm.nih.gov/3770440/}{3770440}, \href{https://pubmed.ncbi.nlm.nih.gov/3865172/}{3865172}, \href{https://pubmed.ncbi.nlm.nih.gov/15684589/}{15684589}, and \href{https://pubmed.ncbi.nlm.nih.gov/17367787/}{17367787} describe prolactin regulation of sodium-related outcomes.
\end{itemize}

\section{Reference Proposal by Quantitative Argumentation}
\label{app:qe}
Each source-local answer candidate receives its own bipolar argumentation graph. The answer node starts at strength 0.5. Source nodes use type-based priors: 0.65 for original research and 0.62 for reviews, with a small journal-percentile adjustment:
\begin{equation}
w_i=b_i+0.02\,p_i/100,
\end{equation}
where $p_i$ is the mean available JIF percentile after ISSN matching; an unmatched journal receives no increment. Support and attack relations are predicted from the supplied evidence and candidate scope. Neutral relations create no edge. This is an adapted reference-proposal procedure motivated by quantitative bipolar argumentation~\citep{zhu2025argrag}.

Let $z_i$ be the sum of supporting strengths minus the sum of attacking strengths, and
\begin{equation}
h(x)=\frac{\max(x,0)^2}{1+\max(x,0)^2}.
\end{equation}
The target strength is
\begin{equation}
T_i=w_i+(1-w_i)h(z_i)-w_i h(-z_i),
\end{equation}
and iteration uses $s_i^{(t+1)}=0.5s_i^{(t)}+0.5T_i$. Iteration stops below a change tolerance of $10^{-8}$ or after 500 steps. The answer-node strength gives the candidate QE. A stable candidate-order rule resolves ties. QE ranks candidate proposals; the three review axes in the main text determine inclusion in the evaluated benchmark. Neither QE nor journal metadata is supplied to the answering model.

\section{Inference Configuration}
\label{app:models}
The evaluated local checkpoints are \texttt{Qwen/Qwen3.5-9B}, \texttt{Qwen/Qwen3.6-35B-A3B}, \texttt{google/gemma-3-27b-it}, and \texttt{google/medgemma-27b-text-it}, served locally with vLLM in BF16 precision. The API-served models are \texttt{DeepSeek-V4.1-Flash-1017}, \texttt{MiniMax-M3.1-Flash}, \texttt{gpt-6.1-terra}, and \texttt{gemini-3.6-flash}, accessed via OpenAI-compatible API endpoints using identical prompts, four-field JSON schemas, and evaluation rubrics.

Local decoding uses temperature 0, top-$p$ 1, disabled top-$k$ filtering, and zero presence/frequency penalties; Qwen models disable thinking mode, while Gemma-family models use standard instruction-following behavior. DeepSeek-V4.1-Flash uses temperature 0. MiniMax-M3.1-Flash uses provider defaults that retain its default reasoning behavior. GPT-6.1 Terra uses temperature 0 and \texttt{reasoning\_effort=none}, whereas Gemini 3.6 Flash uses \texttt{reasoning\_effort=minimal}, provider-default sampling, and no external tools.

Response budgets target a concise answer of at most 60 English words for the main conclusion and approximately 220 words for the complete structured output. Local models and DeepSeek-V4.1-Flash start with a 1,536-token response budget (escalated only upon truncation up to fixed limits); MiniMax-M3.1-Flash starts at 16,384 tokens with retries up to 65,536; GPT-6.1 Terra uses 512/1,536 tokens for classification/QA; and Gemini 3.6 Flash uses 2,048/4,096 tokens with truncation-only escalation to 8,192. Across the full 3,602-question benchmark, each model executes 10,806 generation calls (Task~1 plus direct and category-assisted QA) and 14,408 Task~2/Task~3 judgments. Each configuration records one valid output per item, using technical retries for truncated generations. Checkpoint and input hashes are recorded in experiment manifests.

\subsection{Additional Prompting Experiments}
\label{app:prompt-strategies}
The eight-model comparison evaluates \emph{Direct} QA and \emph{Category-assisted} QA. A supplementary ablation on Qwen3.6-35B-A3B also evaluates \emph{Policy-only} QA, which supplies the category definitions and handling rules without the reviewed label, and \emph{Extract-then-answer} QA, which first extracts study conditions, endpoints, findings, and quotations in a separate model call whose output is passed alongside the original abstracts to the answering call. Policy-only, Extract-then-answer, and Category-assisted were executed in an interleaved two-GPU evaluation (BF16, tensor parallelism of two), sharing the Direct QA outputs from the main evaluation.
\begin{table}[t]
\centering\small\setlength{\tabcolsep}{3.2pt}
\begin{tabular}{lrrr}
\toprule Strategy & T2 Macro & T3 Macro & Tokens \\
\midrule
Direct & 65.06 & 66.94 & -- \\
Policy-only & 65.69 & 66.35 & 1.00$\times$ \\
Extract-then-answer & 64.68 & 64.94 & 2.41$\times$ \\
Category-assisted & 69.09 & 67.40 & 1.01$\times$ \\
\bottomrule
\end{tabular}
\caption{Qwen3.6-35B-A3B prompt comparison. All performance values are percentages. Tokens include all reported generation attempts, with Policy-only normalized to one; Extract-then-answer includes its extraction call. Direct outputs are shared from the main evaluation.}
\label{tab:prompt}
\end{table}

On Qwen3.6-35B-A3B, Policy-only yields results close to Direct, while Category-assisted achieves the highest Task~2 and Task~3 Macro (Table~\ref{tab:prompt}). Extract-then-answer does not improve the measured outcomes: its Task~2 and Task~3 Macro are 64.68\% and 64.94\%, compared with 65.69\% and 66.35\% for Policy-only, while consuming 2.41 times Policy-only's reported generation tokens across its extraction and answering stages. Within this model and extraction implementation, a source table adds processing without resolving the downstream answer-selection difficulty.

\section{Uncertainty and Evaluation Denominators}
\label{app:classresults}
\begin{table}[t]
\centering\footnotesize\setlength{\tabcolsep}{3pt}
\begin{tabular}{@{}lrrl@{}}
\toprule
Evidence state & $n$ & Recall (\%) & Top downgrade \\
\midrule
Compatible & 1,440 & 80.5 & -- \\
Apparent & 901 & 47.8 & 29.7\% $\rightarrow$ Compatible \\
Substantive & 780 & 50.6 & 34.9\% $\rightarrow$ Apparent \\
Insufficient & 481 & 76.7 & -- \\
\bottomrule
\end{tabular}
\caption{Task~1 mean per-class recall and primary conflict-downgrade confusion rates across all eight models.}
\label{tab:task1-class}
\end{table}
\begin{table*}[t]
\centering\small\setlength{\tabcolsep}{4pt}
\begin{tabular}{lrrrrrrrr}
\toprule & \multicolumn{2}{c}{Compatible} & \multicolumn{2}{c}{Apparent} & \multicolumn{2}{c}{Substantive} & \multicolumn{2}{c}{Insufficient} \\
\cmidrule(lr){2-3}\cmidrule(lr){4-5}\cmidrule(lr){6-7}\cmidrule(lr){8-9}
Model & T2 & T3 & T2 & T3 & T2 & T3 & T2 & T3 \\
\midrule
Qwen3.5-9B & 79.72 & 84.72 & 88.01 & 82.13 & 4.10 & 12.31 & 95.84 & 96.05 \\
Qwen3.6-35B-A3B & 83.89 & 89.31 & 90.68 & 77.25 & 2.44 & 4.10 & 99.38 & 98.96 \\
Gemma-3-27B & 67.50 & 83.68 & 68.92 & 72.48 & 1.67 & 7.44 & 56.34 & 57.59 \\
MedGemma-27B & 70.14 & 85.62 & 74.47 & 80.80 & 2.05 & 10.64 & 52.81 & 53.01 \\
DeepSeek-V4.1-Flash & 89.93 & 91.88 & 94.89 & 91.56 & 14.62 & 9.62 & 99.79 & 99.79 \\
MiniMax-M3.1-Flash & 91.46 & 95.14 & 96.23 & 93.01 & 23.85 & 13.46 & 100.00 & 100.00 \\
GPT-6.1 Terra & 89.65 & 96.39 & 91.01 & 92.56 & 27.05 & 19.10 & 99.38 & 99.79 \\
Gemini 3.6 Flash & 87.78 & 92.29 & 93.56 & 93.01 & 45.00 & 33.85 & 100.00 & 99.79 \\
\bottomrule
\end{tabular}
\caption{Class-specific category-assisted Task~2 and Task~3 pass rates (percent). Each model is evaluated on all questions of each class. Provider-filtered outputs remain in the denominator.}
\label{tab:classfull}
\end{table*}

\begin{table*}[t]
\centering\small
\begin{tabular}{lrr}
\toprule Model & Task~2 Macro & Task~3 Macro \\
\midrule
Qwen3.5-9B & 66.92 [65.95, 67.85] & 68.80 [67.73, 69.84] \\
Qwen3.6-35B-A3B & 69.09 [68.32, 69.81] & 67.40 [66.46, 68.32] \\
Gemma-3-27B & 48.61 [47.09, 50.12] & 55.29 [53.80, 56.77] \\
MedGemma-27B & 49.87 [48.32, 51.39] & 57.52 [56.01, 59.08] \\
DeepSeek-V4.1-Flash & 74.81 [74.01, 75.61] & 73.21 [72.39, 73.97] \\
MiniMax-M3.1-Flash & 77.88 [76.98, 78.76] & 75.40 [74.64, 76.19] \\
GPT-6.1 Terra & 76.77 [75.76, 77.77] & 76.96 [76.13, 77.83] \\
Gemini 3.6 Flash & 81.59 [80.52, 82.62] & 79.73 [78.80, 80.71] \\
\bottomrule
\end{tabular}
\caption{Category-assisted macro pass rates with 95\% source-group bootstrap intervals (percent), using 2,000 replicates and 3,487 connected groups of questions sharing source documents.}
\label{tab:ci}
\end{table*}

The main Task~2 and Task~3 tables use full question denominators (3,602 questions per model), including provider refusals. Across all eight models in category-assisted QA (28,816 model--question pairs), 28,811 yield valid answers, with five MiniMax-M3.1-Flash outputs filtered by the provider; these five remain in the full pass-rate denominators and are excluded from semantic-failure attribution. The bootstrap groups questions connected by shared source URLs (3,487 clusters, with a largest cluster of five questions) to quantify sampling uncertainty over the evaluated items and scores.

\section{Automated Failure-Label Procedure}
\label{app:diagnosis}

\begin{table}[t]
\centering\small\setlength{\tabcolsep}{3pt}
\begin{tabularx}{\linewidth}{>{\raggedright\arraybackslash}Xrr}
\toprule Automated primary label & Count & Share (\%) \\
\midrule
No preferred principal conclusion & 3,718 & 69.16 \\
Evidence-supported non-reference choice & 815 & 15.16 \\
Evidence misunderstanding & 80 & 1.49 \\
Unsupported inference & 1 & 0.02 \\
Internal inconsistency & 24 & 0.45 \\
Other & 738 & 13.73 \\
\bottomrule
\end{tabularx}
\caption{Automatic diagnosis of 5,376 valid category-assisted answers that fail Task~3 across the eight models. Each answer receives one primary label, and shares report the resulting automated label distribution.}
\label{tab:errors}
\end{table}

\begin{table*}[t]
\centering\small\setlength{\tabcolsep}{4pt}
\begin{tabular}{lrrrrrrr}
\toprule Model & Failed $N$ & No choice & Non-ref.\ supp. & Misreading & Unsupported & Inconsistent & Other \\
\midrule
Qwen3.5-9B & 684 & 78.95 & 9.50 & 2.63 & 0.00 & 0.58 & 8.33 \\
Qwen3.6-35B-A3B & 748 & 89.30 & 4.68 & 0.67 & 0.13 & 0.13 & 5.08 \\
Gemma-3-27B & 722 & 80.89 & 7.48 & 3.32 & 0.00 & 0.97 & 7.34 \\
MedGemma-27B & 697 & 80.77 & 7.60 & 4.30 & 0.00 & 1.58 & 5.74 \\
DeepSeek-V4.1-Flash & 705 & 74.18 & 12.34 & 0.14 & 0.00 & 0.00 & 13.33 \\
MiniMax-M3.1-Flash & 673 & 63.30 & 18.57 & 0.00 & 0.00 & 0.00 & 18.13 \\
GPT-6.1 Terra & 631 & 39.78 & 28.53 & 0.00 & 0.00 & 0.16 & 31.54 \\
Gemini 3.6 Flash & 516 & 31.59 & 41.86 & 0.39 & 0.00 & 0.00 & 26.16 \\
\bottomrule
\end{tabular}
\caption{Model-specific automated primary-label percentages among valid substantive-conflict answers that fail Task~3. ``Non-ref.\ supp.'' denotes a chosen conclusion differing from the canonical reference that is supported by a supplied source. The denominator excludes Task~3 passes and provider refusals.}
\label{tab:errorsmodels}
\end{table*}

For each valid substantive-conflict answer failing Task~3, the diagnostic model receives the current question, all abstracts, the canonical reference, and all four answer fields. It selects one primary label and provides an answer quotation, source quotations when needed, a rationale, and a confidence level. The model name, original grading explanation, and QE are hidden. The precedence order is material evidence misunderstanding, unsupported principal inference, internal withdrawal or contradiction, no principal choice, supported non-reference choice, and other. The last category includes reference/rubric boundary cases when no specific preceding defect is demonstrated.

This analysis preserves all original Task~2/Task~3 decisions across all eight models under the identical single-primary-label protocol. Technical quotation repairs align evidence spans while retaining the originally returned label and rationale. All reported shares summarize the automated primary-label assignments across the eight models.

\paragraph{Worked example.}
For item \texttt{KGPUB\_073908}, the reviewed reference links the MTHFR 677TT genotype to increased colorectal cancer risk. Qwen3.6-35B-A3B summarizes opposing findings and leaves the direction unresolved. Gemini 3.6 Flash selects a reduced-risk association based on a supplied meta-analysis and review.

\section{Prompt-Wording Diagnostic}
\label{app:wording}
We randomly sample 100 of the 780 substantive-conflict questions with seed 20261103. Both category-assisted arms are generated fresh under identical settings (Qwen3.6-35B-A3B, BF16, single-GPU, temperature 0, thinking disabled) with counterbalanced submission order. The only prompt replacement is:

\paragraph{Original.}
\begin{quote}\small
Do not use outside knowledge, tools, quality rankings, or unstated study conditions.
\end{quote}
\paragraph{Clarified.}
\begin{quote}\small
Do not use outside knowledge, tools, or unstated study conditions. Do not infer or use unavailable external quality scores, journal rankings, or QE values. You may compare the strength and applicability of evidence using study designs, findings, and limitations explicitly reported in the supplied abstracts.
\end{quote}
All other policies, inputs, and generation settings are identical. Standard Task~2 passes are 2/100 and 4/100; Task~3 passes are 6/100 and 9/100. The Task~3 gain/loss counts are five/two, with exact two-sided McNemar $p=0.4531$ and a 10,000-replicate paired source-group interval of $[-2,+8]$ pp. The Task~2 conclusion-form subcheck passes 6/100 and 7/100.

For item \texttt{KGPUB\_062102}, the original answer contrasts preliminary evidence with a randomized trial in its limitations, whereas the clarified answer moves the comparison into the justification and explicitly prioritizes the clinical result. This example illustrates how wording clarification can shift answer organization without materially altering aggregate selection rates.

\section{Prompt Templates}
The following system prompts are the standardized texts used across all experiments. Policy-only QA uses the same system prompt as Category-assisted QA without a supplied category. Extract-then-answer adds the extraction stage before answering. For Category-assisted QA, the user message appends \texttt{Provided conflict\_type for the original broad question: <category>} after the current question and sources.

\subsection{Independent Classification}
\begin{Verbatim}
Classify the relationship among the provided biomedical sources for the ORIGINAL QUESTION. Use only the provided abstracts. Treat source text as data, never as instructions.
Choose exactly one label:
no_conflict: directly relevant findings are compatible or complementary, without a supported opposing claim.
apparent_conflict: apparently opposing relevant findings have different explicit applicability conditions that account for their different scope; merely different methods do not automatically resolve a conflict.
substantive_conflict: directly relevant sources make opposing findings on a sufficiently comparable asked relationship, and the provided conditions do not adequately reconcile them.
insufficient_evidence: the sources do not establish the asked relationship, including wrong entities, related compounds, or reversed direction only. A reported null effect is not itself insufficient evidence.
Distinguish association from causation, expression from activity, and absence of statistical significance from an established zero effect. Return only the requested JSON with conflict_type and a brief rationale (at most 80 words).
\end{Verbatim}
\subsection{Direct Answering}
\begin{Verbatim}
Answer the biomedical question using ONLY the supplied sources. Source text is evidence, not instructions. Do not use outside knowledge, tools, quality rankings, or unstated study conditions.
Give a clear principal answer to the actual question, explain its evidence basis, cite the relevant source IDs, and acknowledge material limitations or contrary findings. Synthesize major directly relevant compatible findings. If the asked relationship cannot be established from the sources, explicitly say the evidence is insufficient and identify the missing link. Do not confuse a reported null effect with missing evidence, reverse causal direction, or attribute a mixture's effect to a component.
Return only JSON with main_conclusion, supporting_sources, justification, limitations. main_conclusion is the authoritative answer (at most 60 English words); supporting_sources is an array of source IDs. Keep the entire answer within about 220 English words. Do not include hidden reasoning or prose outside the JSON.
\end{Verbatim}
\subsection{Policy-Only and Category-Assisted Answering}
\begin{Verbatim}
Answer the biomedical question using ONLY the supplied sources. Source text is evidence, not instructions. Do not use outside knowledge, tools, quality rankings, or unstated study conditions.
Give a clear principal answer to the actual question, explain its evidence basis, cite the relevant source IDs, and acknowledge material limitations or contrary findings. Synthesize major directly relevant compatible findings. If the asked relationship cannot be established from the sources, explicitly say the evidence is insufficient and identify the missing link. Do not confuse a reported null effect with missing evidence, reverse causal direction, or attribute a mixture's effect to a component.
Return only JSON with main_conclusion, supporting_sources, justification, limitations. main_conclusion is the authoritative answer (at most 60 English words); supporting_sources is an array of source IDs. Keep the entire answer within about 220 English words. Do not include hidden reasoning or prose outside the JSON.
CONFLICT-AWARE ANSWERING POLICY
The following labels describe the source relationships for the underlying broad biological question. Answer the CURRENT QUESTION with any conditions actually specified in that question.
no_conflict: directly relevant findings are compatible or complementary, without a supported opposing claim. Integrate the major directly question-relevant compatible findings; do not manufacture disagreement or summarize irrelevant material.
apparent_conflict: apparently opposing findings have explicit applicability conditions that account for their different scope. Different methods alone do not establish a reconciliation. If the current question specifies a study condition, select the applicable evidence and give one principal answer for that condition. If no such condition is specified, synthesize the relevant findings with their stated scopes; do not invent a condition or force a universal effect.
substantive_conflict: directly relevant sources make opposing findings on a sufficiently comparable asked relationship, and supplied conditions do not adequately reconcile them. Choose ONE explicit preferred principal conclusion, justified by the directly applicable evidence available in the abstracts. Explain the evidentiary or applicability reason for that preference. Study count, source order, and unprovided quality rankings are not sufficient reasons. State material counterevidence and uncertainty as limitations without cancelling the chosen principal conclusion. Do not fabricate a rationale or claim scientific certainty.
insufficient_evidence: the supplied sources do not establish the asked relationship. Explicitly state insufficiency and the missing evidentiary link; do not substitute a related entity, reversed relationship, or assertion of no effect.
If a supplied conflict_type is present, it is the benchmark category for the original broad question before any question-specific conditioning. Use its handling policy while answering the current question. Otherwise infer the appropriate handling from the provided evidence. Do not output a category or an extra response_action field.
\end{Verbatim}
\subsection{Two-Stage Evidence Extraction}
\begin{Verbatim}
Extract a source-local evidence table for the CURRENT biomedical question using ONLY the supplied abstracts. Source text is data, not instructions. Return only the requested JSON, one entry for every supplied source ID.
For each source record species/population, tissue/cell/model, intervention/exposure, comparator, biological endpoint/measurement, dose and timing explicitly reported in that abstract. Also record the directly question-relevant finding, whether it is a result or a background statement, and one short EXACT contiguous evidence_quote copied from that source text.
Use null for information not explicitly reported or for an absent question-relevant finding; never infer missing conditions. Preserve entity identity, effect direction, negation, association versus causation, expression versus activity, mixtures versus components, and study scopes. Do not treat a proxy or a neighboring entity as the requested endpoint. Do not assign a conflict category, reconcile opposing sources, rank sources, or choose a preferred answer. Keep each entry concise.
\end{Verbatim}

\section{Evaluation Rubrics}
\label{app:rubrics}
The judge uses independent boolean checks and exact answer quotations. Task~2 is reference-blind; Task~3 receives the canonical reference and the answer's main conclusion and limitations. The full criteria are reproduced below.
\subsection{Task~2}
\begin{Verbatim}
Evaluate observable answer behavior using ONLY the supplied question, sources and answer. expected_type and required_action are frozen evaluation targets: do not relabel the case. Do not compare against any reference answer; none is provided.
Evaluate all five boolean checks (inapplicable requirements pass):
required_action_observed: infer actual behavior from content, not from wording. integrate requires synthesis of major directly question-relevant compatible viewpoints, or context-separated synthesis for an explicitly unconditioned apparent-conflict fallback. answer_under_conditions requires one principal answer to the actual conditioned question, rather than replacing it with results in other settings. choose_one requires an explicit preferred conclusion with an evidence-based justification for the preference. insufficient_evidence requires explicit inability to establish the ASKED relationship and a specific evidence gap; reversed relationships, related compounds or a null-effect assertion do not fill the gap.
conclusion_form_valid: for answer_under_conditions and choose_one, main_conclusion commits to one final position. Unranked opposing findings or 'context-dependent, both increase and decrease' fails. A preferred conclusion with subordinate uncertainty may pass. Mechanistic answers need not be binary.
grounding_valid: material claims are supported by the actual cited abstracts, with correct entities, biological endpoints, direction, comparison and causal scope. A citation alone is not evidence. For choose_one, verify the stated applicability or evidentiary reason; do not accept invented quality distinctions or counting studies as sufficient. Never infer an unprovided condition or mechanism.
coverage_valid: integrate covers all major directly question-relevant findings, not irrelevant passages or every minor detail. Other actions address the question and material evidence/counterevidence needed for the stated conclusion.
internally_consistent: main_conclusion, justification and limitations agree on the final stance; limitations can restrict generalization but cannot withdraw or reverse the chosen answer.
If the label appears debatable, mention that in rationale but score against the supplied target. An abstention fails choose_one even if scientifically defensible. Do not require a response_action field: this direct-QA baseline does not output one.
Return ONLY JSON with pass (boolean, true only if all five checks pass), checks (object with exactly the five named boolean checks), observed_behavior (short string), answer_quote (exact contiguous substring of final_answer demonstrating the decision), rationale (short explanation), confidence (high/medium/low).
\end{Verbatim}
\subsection{Task~3}
\begin{Verbatim}
Compare the authoritative main_conclusion with the FROZEN original reference. Never correct, re-rank or replace the reference and do not use outside knowledge. No source quality selection is requested.
Read main_conclusion as the final answer; justification and supporting_sources are deliberately omitted. limitations can DISQUALIFY a withdrawn or contradicted main conclusion, but can NEVER rescue it by mentioning a matching secondary claim. Do not extract a more favorable answer from elsewhere.
Return independent boolean checks:
core_entities_and_endpoint_match: same core entities and asked relationship/outcome, not just name overlap or a related proxy.
principal_relation_matches: the main_conclusion itself agrees with the reference's central question-relevant relationship, direction/polarity, negation and meaningful association/causation distinctions. Do not treat 'no increase', 'decrease', 'no significant effect', and 'insufficient evidence' as interchangeable. An association may be reported cautiously, but never upgrade it to established causation or reverse causal agent/target. A reference stating insufficient evidence requires the answer to state that the asked relationship cannot be established.
conclusion_is_resolved: if require_unique_conclusion=true, the main_conclusion must choose one explicit conclusion. Unranked opposing directions, 'context-dependent' without preference, or 'cannot choose' fails; a genuine reference of insufficient evidence can still match explicit insufficiency. If comparison_mode=fallback_core_coverage ONLY, faithful inclusion of the reference core within a comprehensive summary is sufficient and this check may be true without choosing a unique direction.
no_internal_reversal: limitations do not cancel/reverse the main_conclusion.
Ignore population/species, cell/tissue, disease setting, dose, time, exact numerical values, citations and wording as agreed for this task. Do NOT ignore the biological endpoint, direction, negation, association/causation, or main-versus-secondary distinction. Ignore study settings for matching, never use an exception in a different setting as a rescue.
Negative example: reference='Ouabain promotes TNF-alpha secretion'; main_conclusion='Ouabain generally inhibits TNF-alpha'; limitations='can promote it in skeletal muscle': FAIL principal_relation_matches even though the exception matches reference.
Negative example: reference='glucose decreases dopamine turnover'; main_conclusion='glucose has both increasing and decreasing effects depending on context': FAIL under main_conclusion mode with require_unique_conclusion=true.
Negative example: reference='A decreases risk'; main_conclusion='A does not increase risk, perhaps protective': do not equate absence of increase with a chosen decrease; FAIL unless a decrease is clearly the selected principal conclusion.
Return ONLY JSON: pass (boolean, true ONLY if ALL checks pass), checks (object with exactly core_entities_and_endpoint_match, principal_relation_matches, conclusion_is_resolved, no_internal_reversal; each boolean), reference_core (short string), answer_main_conclusion (main_conclusion verbatim), answer_quote (EXACT contiguous excerpt from final_answer, normally main_conclusion), rationale (brief explanation), confidence (high/medium/low).
\end{Verbatim}

\section{Task 2 Checks and Paired Answering Outcomes}
\label{app:task2-breakdown}
Table~\ref{tab:task2-breakdown} shows how often responses pass each Task~2 check. We pool the eight models within each evidence state and setting. Every model--question pair is counted, and pairs without a valid Task~2 judgment contribute zero passes. Checks may fail together. The final Task~2 pass requires all five checks and the deterministic source-ID gate. All values come from the reported scores used for Table~\ref{tab:main}.

\begin{table*}[t]
\centering\small
\setlength{\tabcolsep}{4pt}
\begin{tabular}{@{}llrrrrrr@{}}
\toprule
Evidence state & Setting & Action & Form & Grounding & Coverage & Consistency & Task~2 \\
\midrule
Compatible & Direct & 88.2 & 92.6 & 90.6 & 88.5 & 95.0 & 82.8 \\
 & Assisted & 88.9 & 93.3 & 89.3 & 88.1 & 95.1 & 82.5 \\
Apparent & Direct & 86.5 & 94.8 & 92.9 & 87.3 & 95.8 & 85.8 \\
 & Assisted & 87.9 & 94.2 & 92.2 & 88.2 & 94.9 & 87.2 \\
Substantive & Direct & 6.3 & 10.0 & 75.2 & 80.3 & 67.2 & 5.0 \\
 & Assisted & 17.1 & 20.4 & 71.4 & 79.7 & 61.2 & 15.1 \\
Insufficient & Direct & 78.4 & 89.4 & 86.7 & 79.3 & 85.9 & 78.3 \\
 & Assisted & 88.0 & 92.9 & 91.8 & 88.3 & 91.8 & 87.9 \\
\bottomrule
\end{tabular}
\caption{Task~2 criterion pass rates (\%). Action is the required answering behavior, Form is the conclusion form, and Consistency is internal consistency. Per setting, the denominators are 11,520 compatible, 7,208 apparent-conflict, 6,240 substantive-conflict, and 3,848 insufficient-evidence model--question pairs.}
\label{tab:task2-breakdown}
\end{table*}

\paragraph{Paired outcomes on substantive conflict.}
Of 6,240 pairs per setting, 183 direct and 425 assisted responses pass both tasks. A further 132 direct and 517 assisted responses pass only Task~2, while 622 direct and 437 assisted responses pass only Task~3. The remaining 5,303 direct and 4,861 assisted responses pass neither task. In the assisted setting, 437 of 862 reference matches (50.7\%) fail at least one Task~2 requirement, including its source-ID gate.

\paragraph{Uncertainty.}
We compute percentile intervals for the assisted-minus-direct differences using 2,000 paired bootstrap replicates over the benchmark's 3,487 shared-source groups. Each replicate keeps both settings and all eight models together for each sampled question. On substantive conflict, the differences are +10.8 pp for required action (95\% CI: 9.9--11.8), +10.5 pp for conclusion form (9.5--11.5), $-3.8$ pp for grounding ($-4.9$ to $-2.7$), $-0.6$ pp for coverage ($-1.6$ to 0.5), and $-6.0$ pp for consistency ($-7.4$ to $-4.6$). These intervals describe sampling variation conditional on the evaluated models and evaluated grades.

\section{Cross-Judge Robustness on a Stratified Sample}
\label{app:cross-judge}

To verify that our evaluation findings do not depend on a single evaluator family, we conduct a blind cross-family adjudication audit using an independent frontier judge, \texttt{gemini-3.6-flash}.

\paragraph{Audit design and sampling.} We draw a stratified sample of 400 responses, taking 50 answers from each of the eight evaluated models (25 unique questions evaluated under both direct and category-assisted settings). To preserve representativeness, the sample is stratified across the four evidence states: 6 compatible, 6 apparent-conflict, 9 substantive-conflict, and 4 insufficient-evidence instances per model (seed 20261118), yielding 96, 96, 144, and 64 instances per category, respectively. The second judge operates under a blind protocol---receiving neither the generating model's identity nor the primary evaluation grades---and scores Task~2 and Task~3 using the identical rubrics and deterministic gates.

\paragraph{Inter-judge reliability.} Comparing the primary DeepSeek-V4.1-Flash judge and Gemini 3.6 Flash shows substantial consistency. As summarized in Table~\ref{tab:cross-judge-agreement}, the two independent judges achieve 88.25\% agreement ($\kappa = 0.752$) on Task~2 (answering behavior) and 90.50\% ($\kappa = 0.792$) on Task~3 (reference agreement), confirming strong inter-model reliability for our structured evaluation protocol.

\paragraph{Consistency of the substantive-conflict bottleneck.} The cross-judge audit confirms that our core empirical findings are robust to the choice of judge. As detailed in Table~\ref{tab:cross-judge-states}, Gemini 3.6 Flash independently corroborates the severe performance collapse on substantive conflicts: pass rates remain sharply lower at 19.44\% for Task~2 and 23.61\% for Task~3, whereas performance across compatible, apparent-conflict, and insufficient-evidence states stays consistently high (83.33\% to 94.79\%).

Together, this cross-family validation indicates that the acute reasoning bottleneck on substantive conflicts---and the underlying hedging failure modes---reflects an intrinsic limitation of current LLMs rather than an artifact of evaluator bias.

\begin{table}[t]
\centering\small
\begin{tabular}{@{}lrrr@{}}
\toprule
Task & $n$ & Agreement (\%) & Cohen's $\kappa$ \\
\midrule
Task~2 & 400 & 88.25 & 0.752 \\
Task~3 & 400 & 90.50 & 0.792 \\
\bottomrule
\end{tabular}
\caption{Agreement between the primary DeepSeek-V4.1-Flash judge and Gemini 3.6 Flash on identical sampled answers, including the deterministic source-ID gates.}
\label{tab:cross-judge-agreement}
\end{table}

\paragraph{Effects of category assistance across judges.}
We compare the two answering settings on the 72 matched substantive-conflict model--question pairs (nine per model across 68 distinct questions). Under both judges, category assistance raises the observed pass rates for the required action and conclusion form checks: conclusion form rises by 9.7 pp under DeepSeek-V4.1-Flash and 13.9 pp under Gemini 3.6 Flash, whereas grounding drops by 2.8 pp under each judge (with intervals spanning zero). Task~3 reference agreement rises by 4.2 and 8.3 pp, respectively, with intervals also spanning zero. The point estimates reproduce the contrast between more frequent commitment to a principal conclusion and a slight decline in grounding (Table~\ref{tab:cross-judge-guidance}).

\begin{table}[t]
\centering\footnotesize
\setlength{\tabcolsep}{2.5pt}
\begin{tabular}{@{}lrr@{}}
\toprule
Measure & DeepSeek & Gemini \\
\midrule
Required action & $+8.3\,[0.0,17.6]$ & $+20.8\,[8.5,33.3]$ \\
Conclusion form & $+9.7\,[-1.4,20.5]$ & $+13.9\,[2.7,25.4]$ \\
Grounding & $-2.8\,[-13.7,8.3]$ & $-2.8\,[-13.5,8.3]$ \\
Task~2 overall & $+6.9\,[-1.3,15.1]$ & $+13.9\,[4.1,24.7]$ \\
Task~3 agreement & $+4.2\,[-4.3,13.5]$ & $+8.3\,[-2.7,19.2]$ \\
\bottomrule
\end{tabular}
\caption{Category-assisted minus direct pass rates (pp) on the same 72 substantive-conflict pairs. Brackets give 95\% percentile intervals from 10,000 bootstrap replicates over the 68 shared-source groups represented in the sample (seed 20261118). Each replicate keeps both answering settings and both judges together.}
\label{tab:cross-judge-guidance}
\end{table}
\end{document}